%% file: main.tex
\documentclass[10pt,twocolumn]{article}

\usepackage{fullpage}

\usepackage{graphicx}
\usepackage{multirow}
\usepackage{array}
\usepackage{flushend}
\usepackage{fancyvrb}
\usepackage{times}
\usepackage{amsmath}
\usepackage{amsfonts}
\usepackage{algorithm}
\usepackage{listings}
\usepackage{appendix}
\usepackage{subcaption}
\usepackage[noend]{algpseudocode}
\usepackage[usenames,dvipsnames,svgnames,table]{xcolor}
\usepackage[pdftex, colorlinks=true, citecolor=black, linkcolor=black, citebordercolor={0 0 0}, urlcolor=black]{hyperref}

\input{macros.tex}
\begin{document}

\title{\bf SplitBrain: Hybrid Data and Model Parallel Deep Learning}
\author{
\textbf{Farley Lai\thanks{Work done as a NEC Labs intern in 2016.}, Asim Kadav, Erik Kruus}\\
NEC Laboratories America, Inc., Princeton, NJ, USA\\
\texttt{\{farleylai,asim,kruus\}@nec-labs.com}
}
\date{}
\maketitle
\thispagestyle{empty}
\lstdefinestyle{default}{
  captionpos=b,
  xleftmargin=2.5em,
  resetmargins=true,
  belowcaptionskip=1\baselineskip,
  breaklines=true,
  frame=L,
  numbers=left,
  tabsize=2,
  showstringspaces=false,
  basicstyle=\footnotesize\ttfamily,
}

\input{sections/abstract}
\input{sections/intro}
\input{sections/background}
\input{sections/design}

\input{sections/impl}
\input{sections/evaluation}
\input{sections/related}
\input{sections/conclusion}

{\small
\bibliographystyle{ieee_fullname}
\bibliography{main}
}

\begin{appendices}
\end{appendices}
\end{document}

%% file: macros.tex
\newcommand{\mycomment}[1]{}
\newcommand{\ignore}[1]{}
\ifpdf
\renewcommand{\url}[1]{\texttt{\small{#1}}}
\else
\newcommand{\url}[1]{\texttt{\small{#1}}}
\fi

\newcommand{\codesm}[1]{\texttt{\small #1}}

\newcommand{\toolshort}{SplitBrain}

\makeatletter
\def\BState{\State\hskip-\ALG@thistlm}
\makeatother

%% file: sections/abstract.tex
\begin{abstract}

The recent success of deep learning applications has coincided with those widely available powerful computational resources for training sophisticated machine learning models with huge datasets. 
Nonetheless, training large models such as convolutional neural networks using model parallelism (as opposed to data parallelism) is challenging because the complex nature of communication between model shards makes it difficult to partition the computation efficiently across multiple machines with an acceptable trade-off.
This paper presents \toolshort, a high performance distributed deep learning framework supporting hybrid data and model parallelism.
Specifically, \toolshort~provides layer-specific partitioning that co-locates compute intensive convolutional layers while sharding memory demanding layers.
A novel scalable group communication is proposed to further improve the training throughput with reduced communication overhead.
The results show that \toolshort~can achieve nearly linear speedup while saving up to 67\% of memory consumption for data and model parallel VGG over CIFAR-10.

\end{abstract}


%% file: sections/intro.tex
\section{Introduction}
\label{sec:intro}
Deep learning has seen a resurgence in the last few years owing to their superior performance over unstructured data such as text, images and video.
This has been driven by emerging deep models to train large datasets such as ImageNet \cite{krizhevsky2012imagenet, russakovsky2015imagenet}.
The model training is unprecedentedly immense and inevitably relies on the availability of efficient computation over CPUs, GPUs and communication infrastructures.
Convolutional neural networks (CNNs) based deep models are extensively used in image and video recognition, natural language processing (NLP) and other machine learning (ML) applications.
The success of CNNs in these areas corresponds to a significant increase in the number of parameters and the amount of computation.

To address the massive demand on space and performance, both data parallelism (DP) and model parallelism (MP) are taken into consideration to scale the computation over multiple machines and leverage distributed computing resources.
DP partitions the input dataset $\{x_1,~x_2,~...,~x_n\}$ for networked computing nodes as model replicas to process in parallel.
Each model replica trains input partition locally and only synchronize the full model parameters with one another through a \codesm{reduce} operation every fixed number of iterations.

On the other hand, MP splits the model $\{w_1,~w_2,~...,~w_n\}$ into model shards trained on distributed machines with reduced memory and computational requirements.
Unlike DP which requires machines with comparable computing resources, MP potentially facilitates heterogeneity across computing nodes.
However, providing efficient MP is complicated because feedforward operations and backpropagation for each training sample now involve communication between model shards.
Moreover, increased communication complexity may hinder the throughput significantly.

To make the best of both worlds, we build \toolshort, which supports hybrid DP and MP across multiple machines for efficiently training deep CNNs.
Our contributions to deep learning include:
\begin{itemize}
\item We present \toolshort, an asynchronous approximate computation model for large-scale data and model parallel applications.
\item Automatic layer partitioning and network transformation to hide MP complexity from CNN developers.
\item Hybrid MP group extension to trade off between memory usage and communication efficiency. 
\end{itemize}

In our evaluation, we find that \toolshort\ scales linearly up to a large number of machines, and is configurable to trade off between communication overhead and memory usage. 

The rest of the paper is organized as follows. 
In the next section, we provide the background on CNNs.
Section \ref{sec:design} introduces the hybrid design and extension to trade off the communication overhead.
The implementation is detailed in Section \ref{sec:impl}.
Section \ref{sec:evaluation} evaluates the performance in terms of throughput and memory requirements.
Related work and conclusions are given at the end.


%% file: sections/background.tex
\section{Background}
\label{sec:back}




In this section, we first cover the basics of distributed machine learning and then discuss the computational requirements of CNNs followed by the challenges to scaling out CNN performance. 

\subsection{Distributed Machine Learning}

\begin{figure}
\centering
\includegraphics[keepaspectratio=true,width=3.2in]{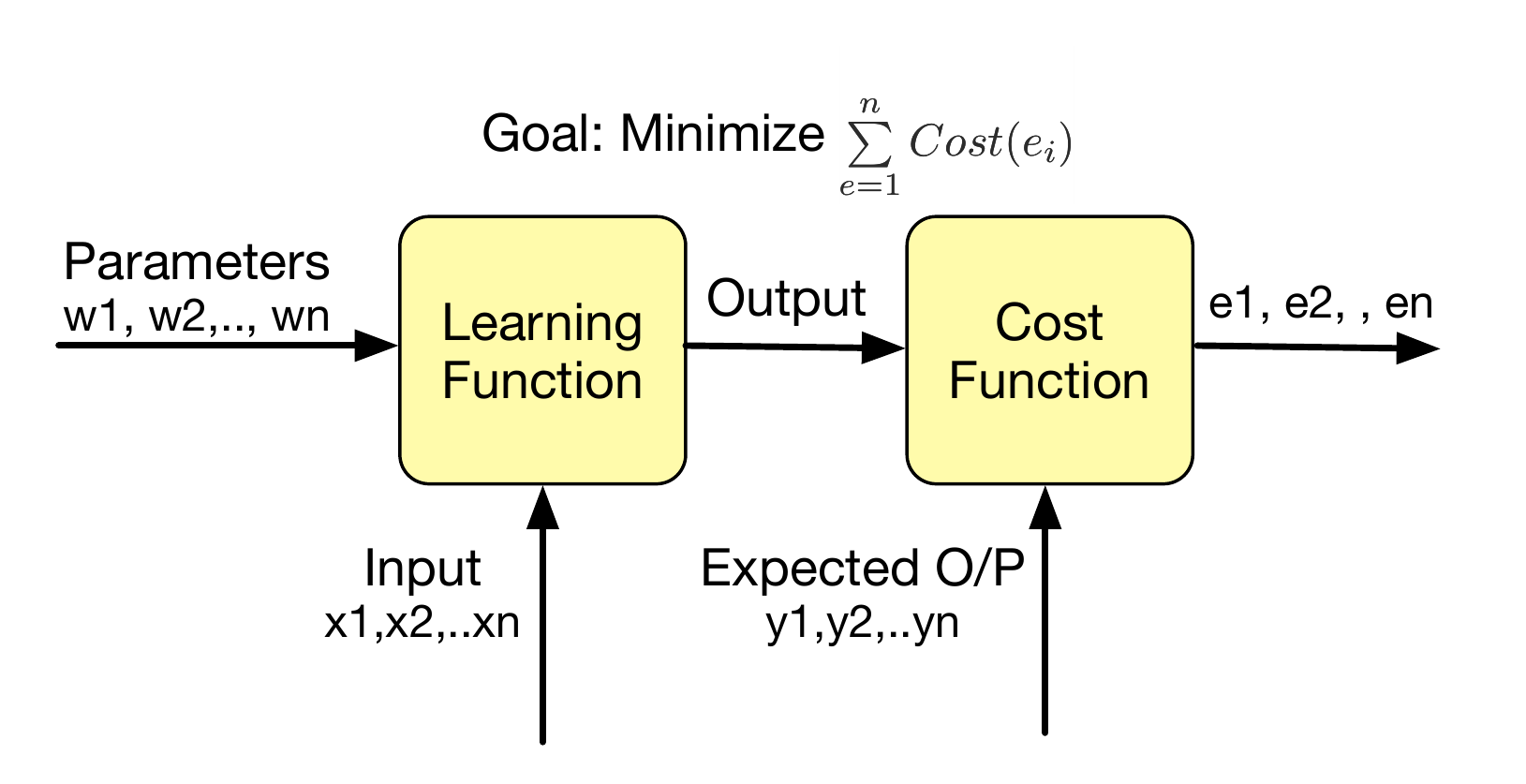}
\caption{\small \bf Machine learning is an iterative training process in search of a set of parameters that minimize the classification error. Both the model parameters $\{w_1,~w_2,~...,~w_n\}$ and an input dataset examples $\{x_1,~x_2,~...,~x_n\}$ can be viewed as vectors. The learning function takes some combination of both to produce the output. The cost function evaluates the error between the output and the expected value. The error are fed back to update to the parameters such that the error is minimized. The training continues until the error converges to an acceptable value.}
\label{fig:ml-problems}
\vspace{-0.2in}
\end{figure}

Machine learning algorithms build a statistical model to train over a dataset and make predictions of unseen data.
The model can be represented as a parameter vector $W$ which is updated iteratively during the training phase.
The training terminates when the model is evaluated against a validation set with an acceptable accuracy.
Originally, it is required to iterate through the entire training set for one time update of the model parameters.
As the dataset grows larger and larger, it is very time-consuming for the model updates to converge.
Consequently, ML practitioners often use the Stochastic Gradient Descent (SGD) to update the parameters every time a randomly selected example or mini-batch is trained.
Figure~\ref{fig:ml-problems} shows SGD can be used to train ML models iteratively over the dataset. 

Unfortunately, even SGD with mini-batching is insufficient for training a huge dataset at the scale of hundreds of millions of examples efficiently.
One way to achieve significant speedup is to incorporate massive parallel computing resources.
To scale out the computation over multiple machines, the SGD can be distributed over a computing cluster in data parallel by splitting the input dataset examples $\{x_1,~x_2,~...,~x_n\}$, and/or in model parallel by splitting the model parameters $\{w_1,~w_2,~...,~w_n\}$. 
Data parallel ML replicates the model on different machines as model replicas that are trained and updated with the partitioned dataset independently.
However, to converge to a global model, the model replicas must be synchronized every fixed number of iterations.
A commonly used synchronization technique is the Bulk Synchronous Parallel (BSP) model where a global barrier ensures that model replicas train and update intermediate parameters at the same speed while agreeing on a global model.
The synchronization involves parameters exchange between model replicas and is followed by a \codesm{reduce} operation to derive the global model.
The exchange of parameters may be carried out in a peer-to-peer fashion or through a centralized parameter server.
A realization of the \codesm{reduce} operation is by averaging the model replicas.

Inevitably, distributed parallel ML suffers from additional synchronization and communication overhead.
Fortunately, these ML algorithms are {\em iterative-convergent} and can tolerate errors in the synchronization step.
Several recent works in \cite{canny:butterfly, cipar2013solving, recht2011hogwild} proposed to reduce the overhead by communicating stale intermediate parameter updates and exchanging parameters with little or no synchronization.
However, performance improvement in terms of throughput by relaxing the synchronization barrier \cite{chilimbi2014adam, wang2013asynchronous} may not always speed up the convergence but lead to an incorrect final value.
This is because the distributed machine workers do not synchronize but communicate model parameters at different speed, the global model may skew in the favor of the workers that train their model parameters faster.
In~\cite{Ho2013-qr}, it is proved that the correctness can be guaranteed despite exploiting the bounded staleness of local cache between workers to speed up the convergence. In ~\cite{Cui2014-eg}, it is shown that the staleness can be tuned to find the sweet spot between performance and convergence.
Furthermore, if a single global model is maintained and updated without locks as in \cite{recht2011hogwild}, a global convergence may only be possible if the parameter vector is sparse.
Finally, maintaining a single global model in a distributed setting results in lots of wasted communication since a lot of the useful parameter updates may be overwritten \cite{chilimbi2014adam, noel2014dogwild}.

The trade-off between training throughput and synchronization overhead remains even with parameter servers.
A parameter server coordinates the parameter consistency amongst the worker machines by resetting their model replicas every iteration and ensures global consensus on the final model.
However, a single server communicating with a large number of workers may result in network congestion.
The distributed parameter server architecture \cite{li2014parameterserver} addresses the problem by limiting network traffic through a central master \cite{Abadi2016-oa, dean2012large, li2014parameterserver}.
Nevertheless, this parameter server architecture still suffers from similar synchronization trade-off as peer-to-peer BSP systems.
Specifically, an asynchronous server may \codesm{reduce} with few workers and slow down the convergence while a synchronous server may spend significant amount of time on the \codesm{barrier} operation.
As a result, it is essential for a distributed ML framework to provide levels of synchronization trade-off in terms of application requirements.


\subsection{Convolutional Neural Networks}

\begin{figure}
\centering
\includegraphics[keepaspectratio=true,width=3.2in]{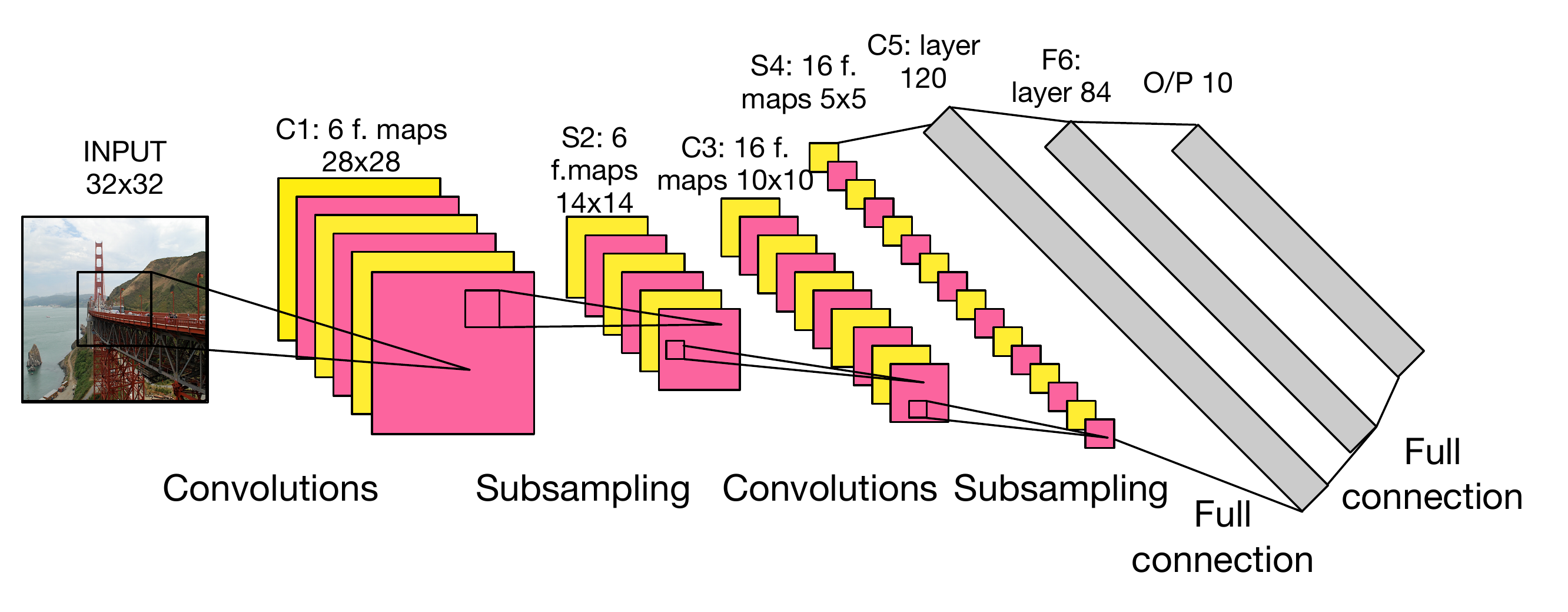}
\caption{\small \bf A representative CNN architecture for image recognition. Features are extracted through convolution layers while Subsampling is done by pooling layers to reduce the computation complexity. FC layers perform the classification at the end.}
\label{fig:cnn}
\vspace{-0.2in}
\end{figure}

Out of numerous ML models, deep CNNs gain success in many application domains such as large-scale image recognition.
We consider training a CNN to classify a large image dataset as shown in Figure \ref{fig:cnn}.
A CNN consists of successive stages or layers of convolution, pooling and non-linear activation functions such as sigmoid or the rectified linear unit (ReLU).
In contrast with traditional ML models that expect extracted features as input, CNNs are constructed with feature extraction in mind and trained in an end-to-end fashion.
In each convolutional stage, successive patches of the input image are multiplied with filters using a convolution operation. 
The filter kernel is typically a square patch of weights used to extract specific features such as edges in an image.
The following pooling layer downsamples the features in preparation for a non-linear activation function.
The combination of convolution, pooling and activation layers may be stacked to extract higher level features from edges, circles to faces.
Eventually, the resulting activations of features go through fully-connected (FC) layers for classification.
The above dataflow is called forward propagation (fprop) as in traditional neural networks (NNs).
During training, the classification errors are backward propagated (bprop) to compute the gradients and update layer parameters.

Compared with NNs, what makes CNNs superior are learnable filters through convolutional layers.
Nonetheless, the convolution operation inevitably introduces intensive computational requirements.
The general trend since past few years has been that the networks have grown deeper with an overall increase in the number of parameters and convolution operations.
These highly capable CNNs tend to impose stringent training and inference cost beyond the capacity of commodity devices.
In particular, it is possible to create a CNN of more than one hundred layers and even billions of parameters but the tremendous space requirements and massive computation make DP impractical with full model replicas.
To overcome the scalability issue, MP is the complementary technique to DP in the sense that the required space and computation of a model are distributed over multiple machines as model shards.
This implies the training and validation of a single example must involve frequent communication between the model shards.
Moreover, a certain layer may contain too large a number of parameters to train on a single machine.
The ML framework must support splitting the model by layer and hide the communication complexity from model builders for ease of development while trading off communication overhead effectively.
We discus how the design of SplitBrain addresses the challenges to combining DP and MP in the next section.

%% file: sections/design.tex
\section{Design}
\label{sec:design}




\toolshort is a ML framework which facilitates the construction of CNNs by allowing programmers to connect parametrized CNN layers.
The three layer types exposed to programmers for ease of building CNNs are convolutional, FC and functional layers.
While convolutional and FC layers are straightforward to specify the kernel and output sizes, functional layers provide pooling, dropout and more to realize functional mappings from input to output.
The design goal is to hide the communication complexity introduced by DP and MP from programmers.
The programmer may configure how the resulting CNN is partitioned across machines but the construction of the CNN remains the same as building a complete local model.
SplitBrain automatically transforms the given CNN into a data and model parallel counterpart by splitting selected layers as well as inserting supporting communication layers.
The procedure is detailed in the following subsections.

\subsection{Model Parallelism}



Inspired by the parallelizing trick in~\cite{krizhevsky2014one} where hybrid CNNs mix DP with MP in general, SplitBrain extends one of the three proposed MP communication schemes for scalability, which was not evaluated originally.
In summary, the trick is based on the observation that convolutional layers tend to be compute intensive with few parameters while FC layers consume the most memory for storing a huge number of parameters.
On one hand, the communication overhead for data parallelizing convolutional layers should be affordable.
On the other hand, model parallelizing FC layers would lead to significant memory savings.
This justifies the practicality of hybrid parallelization of CNNs.
Now, assuming the mini-batch size is $B$ and there are $K$ worker machines, the three proposed schemes in \cite{krizhevsky2014one} are denoted by BK, B and B/K.
Before the FC layers, each worker is processing its own batch in data parallel.
In the boundary with the first FC layer to partition, scheme BK aggregates the batches from the $K$ workers by broadcasting to one another.
For scheme B, the workers take turns to broadcast their own batches.
Scheme B/K allows for each worker to broadcast $B/K$ examples simultaneously for $K$ iterations of fprop and bprop through the FC layers.
Clearly, scheme BK requires the most memory space to store $BK$ incoming examples from the other machines and is unlikely to scale for a large $K$.
Scheme B also suffers from the scaling limitation but it is straightforward to implement and already evaluated in \cite{krizhevsky2014one}.
Scheme B/K scales out of the three at this point and serves as the basis of SplitBrain.
Nonetheless, it is not necessarily the case considering the communication overhead through the FC layers.
This is because model parallelized FC layers must communicate with each model shard in exchange of the output and gradients in case of loss of information.
To address the scalability concern, SplitBrain further introduces the group MP extension to trade off between memory requirements and communication overhead.
The intuition is to localize the communication impact of MP on performance to a subset of worker machines.
In the following, we describe the layer partitioning that transforms the CNN and the two supporting modulo and shard layers, that encapsulate the scheduling and communication complexity absent in the local model.

\paragraph {Layer Partitioning}

\begin{figure*} 
    \begin{minipage}[t]{0.3\linewidth}
        \includegraphics[keepaspectratio,width=\textwidth]{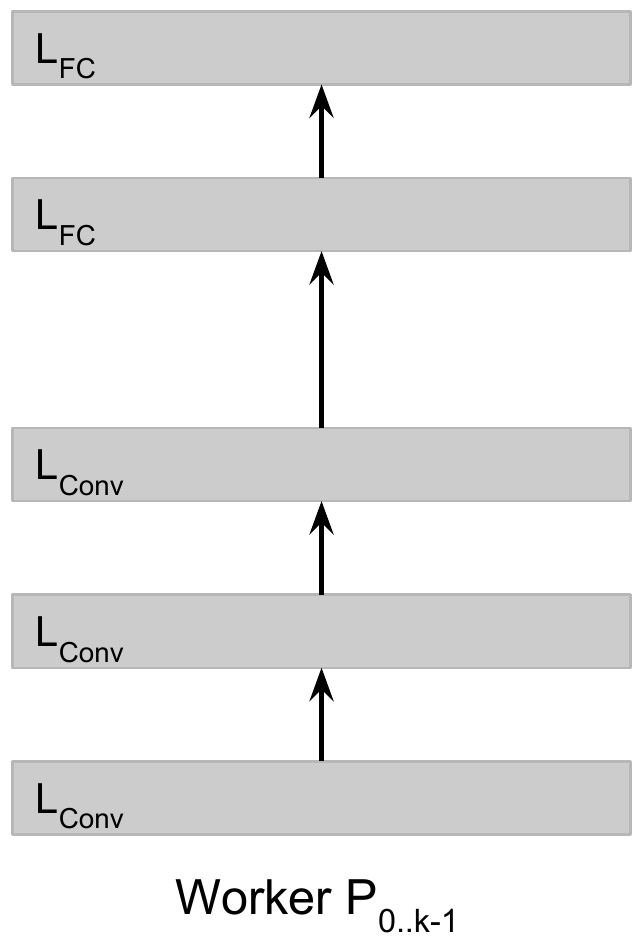}
        \subcaption{\small\bf Original CNN for each worker.}
        \label{fig:cnn-hybrid-a}
    \end{minipage}
    \hfill
    \begin{minipage}[t]{0.65\linewidth}
        \includegraphics[keepaspectratio,width=\textwidth]{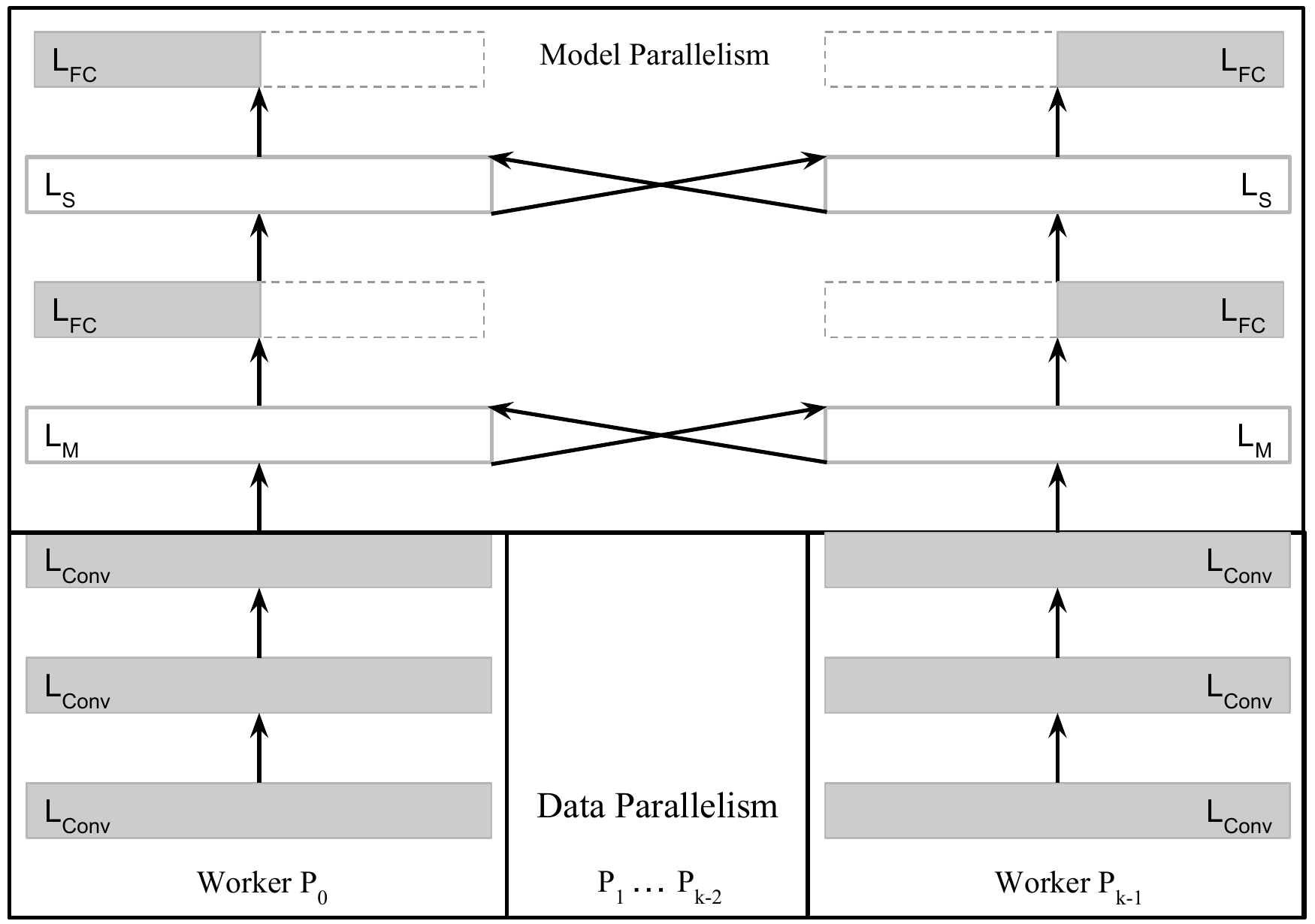}
        \subcaption{\small\bf Transformed CNN with partitioned FC and inserted modulo/shard layers.}
        \label{fig:cnn-hybrid-b}
    \end{minipage}
    \caption{\small\bf Automatic CNN transformation with layer partitioning. 
            $\textrm{L}_\textrm{FC}$, $\textrm{L}_\textrm{Conv}$, $\textrm{L}_\textrm{M}$ and $\textrm{L}_\textrm{S}$ stand for FC, convolutional, modulo and shard layers.}
    \label{fig:cnn-hybrid}
    \vspace{-0.2in}
\end{figure*}

Unlike DP, MP partitions layers to train across multiple machines with reduced memory requirements.
However, not all the layers may be partitioned in a consistent way without layer-specific knowledge.
For example, partitioning a padding layer only requires to communicate additional output around the boundary but full partitioned output exchange is necessary for splitting a FC layer.
Moreover, a CNN might be arbitrarily connected with parallel constructions in general.
Currently, \toolshort~supports partitioning CNNs with common functional and FC layers connected in sequential containers.

\begin{lstlisting}[language=java, style=default, escapechar=\#, caption={\bf Layer Partitioning}, label=alg:partitioning]
/**
 * @param layer current layer to partition
 * @param dim  partitioned input dimension
 * @param dimF full input dimension
 * @param net the transformed output CNN
 */
partition(layer, &dim, &dimF, net) {
  switch(layer) {
  case SEQ: #\label{line:seq}#
    for(L: layer)
      partition(L, dim, dimF, net);
    break;
  case RESHAPE | PAD | CONV | POOLING: #\label{line:skip}#
    if(dim == dimF) 
      dim = dimF = layer.resize(dim); #\label{line:resize}#
    else 
      err("Partitioned input unsupported");
    break;
  case DROPOUT | RELU: #\label{line:func_start}#
    layer.resize(dim);
    break;  #\label{line:func_end}#
  case LINEAR: #\label{line:fc}#
    dimF = layer.out_dim();
    if(dim == dimF) {
      if(useMP() && layer.ccr() > CCR()) { #\label{line:modulo_start}#
        net.push(new MODULO(dimF));
        layer = partition(layer);
      } #\label{line:modulo_end}#
    } else { #\label{line:shard_start}#
      net.push(new SHARD(dim, dimF));
      if(layer.ccr() > CCR())
        layer = partition(layer); #\label{line:shard_end}#
    }
    dim = layer.out_dim(); #\label{line:fc_update}#
    break;
  case LOG_SOFTMAX: #\label{line:softmax_start}#
    if(dim != dimF)
      net.push(new SHARD(dim, dimF)); #\label{line:softmax_end}#
  }
  net.push(layer)
}
\end{lstlisting}

\lstset{
  basicstyle=\footnotesize\ttfamily,
  breaklines=no
}

Listing \ref{alg:partitioning} illustrates the partitioning procedure which tracks the partitioned and full input dimensions of the previous layer output while inserting the modulo and shard layers when necessary.
The procedure takes four parameters.
\lstinline$layer$ points to the current layer to partition.
\lstinline$dim$ and \lstinline$dimF$ are references to \lstinline$layer$ input dimensions w/ and w/o the previous layer partitioned.
\lstinline$net$ is the resulting transformed output CNN.
Initially, the CNN is passed as a sequential container layer to partition, matching the case in line \ref{line:seq}.
Each layer in the sequential container are recursively partitioned in order.
\lstinline$dim$ and \lstinline$dimF$ are subject to modification on return to pass down the dimensions of the partitioned layer output.
Reshape, padding, pooling and convolutional layers are excluded from partitioning and simply propagate their output dimensions to the next layer.
This is because they typically involve significantly fewer parameters than FC linear layers. 
The \lstinline$resize()$ method in line \ref{line:resize} takes an input dimension and returns the output dimension of the layer.
Therefore, only ReLU, dropout and FC layers are considered for actual partitioning.
One-to-one mapping functional layers such as dropout and ReLU adapt to the partitioned input dimension of the previous layer output without modulo or shard layers inserted in between as shown in lines \ref{line:func_start}-\ref{line:func_end} while passing down \lstinline$dim$ and \lstinline$dimF$ intact.
In the case of a FC layer (line \ref{line:fc}), a partitioning decision is made only if the computation to communication ratio (CCR) is sufficiently large.
Since MP inevitably introduces additional communication overhead, intensive enough computation is essential to compensate for the cost.
The computational complexity is estimated in terms of the number of operations in the layer fprop and bprop.
If this is the first FC layer to partition for \lstinline$dim == dimF$, MP enabled, and the CCR above some threshold, a modulo layer is inserted before to schedule the broadcast of $B/K$ examples in the $K$ iterations as shown in lines \ref{line:modulo_start}-\ref{line:modulo_end}.
Otherwise, if the input dimension has been partitioned, a shard layer is inserted before to communicate output from the other workers as shown in lines \ref{line:shard_start}-\ref{line:shard_end}.
In either case, the FC layer is partitioned to $1/K$ size in the overloaded \lstinline$partition(layer)$, and \lstinline$dim$ is updated accordingly in line \ref{line:fc_update}.
Finally, a shard layer is inserted before the softmax layer to provide the full model input if partitioned previously in lines \ref{line:softmax_start}-\ref{line:softmax_end}.
This ensures the same output error is evaluated as a complete local model for back propagation.
In the end, partitioned or not, the current layer is always added to the transformed output CNN.
Figure \ref{fig:cnn-hybrid} demonstrates the transformation by partitioning the FC layers $\textrm{L}_\textrm{FC}$ of the original CNN in Figure \ref{fig:cnn-hybrid-a} and inserting modulo layer $\textrm{L}_\textrm{M}$ and shard layer $\textrm{L}_\textrm{S}$ to produce the resulting CNN in Figure \ref{fig:cnn-hybrid-b}.
Only convolutional and FC layers $\textrm{L}_\textrm{Conv}$ and $\textrm{L}_\textrm{FC}$ are shown for simplification.
Each worker $\textrm{P}_\textrm{k}$ processes a mini-batch in data parallel through $\textrm{L}_\textrm{Conv}$ while iteratively processing the $K$ batches in model parallel through $\textrm{L}_\textrm{FC}$.

\begin{figure}
    \begin{minipage}[b]{1.0\linewidth}
        \includegraphics[keepaspectratio,width=\textwidth]{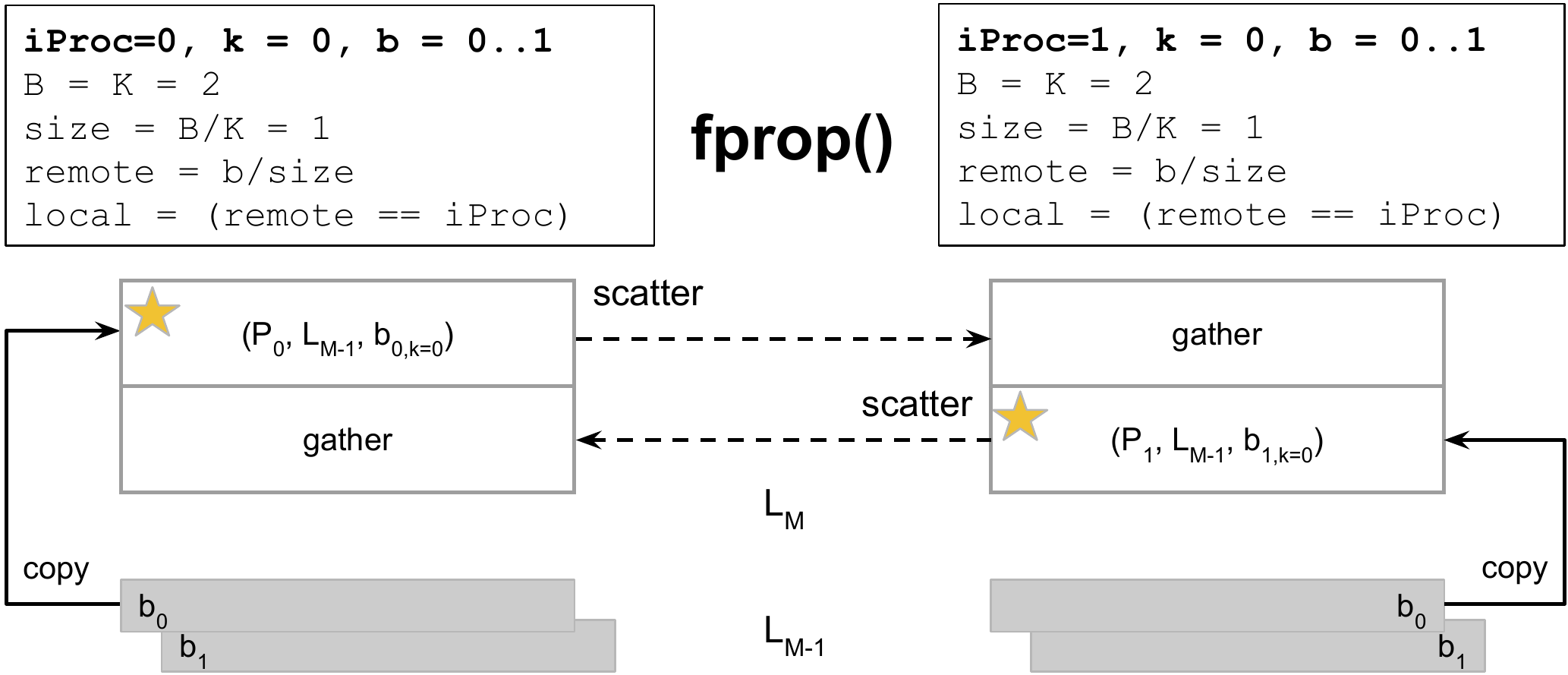}
        \subcaption{Communication of $B/K$ local and $B-B/K$ remote batch examples with the other worker during fprop for $k=0$.}
        \label{fig:modulo-a}
    \end{minipage}
    \hfill
    \begin{minipage}[b]{1.0\linewidth}
        \includegraphics[keepaspectratio,width=\textwidth]{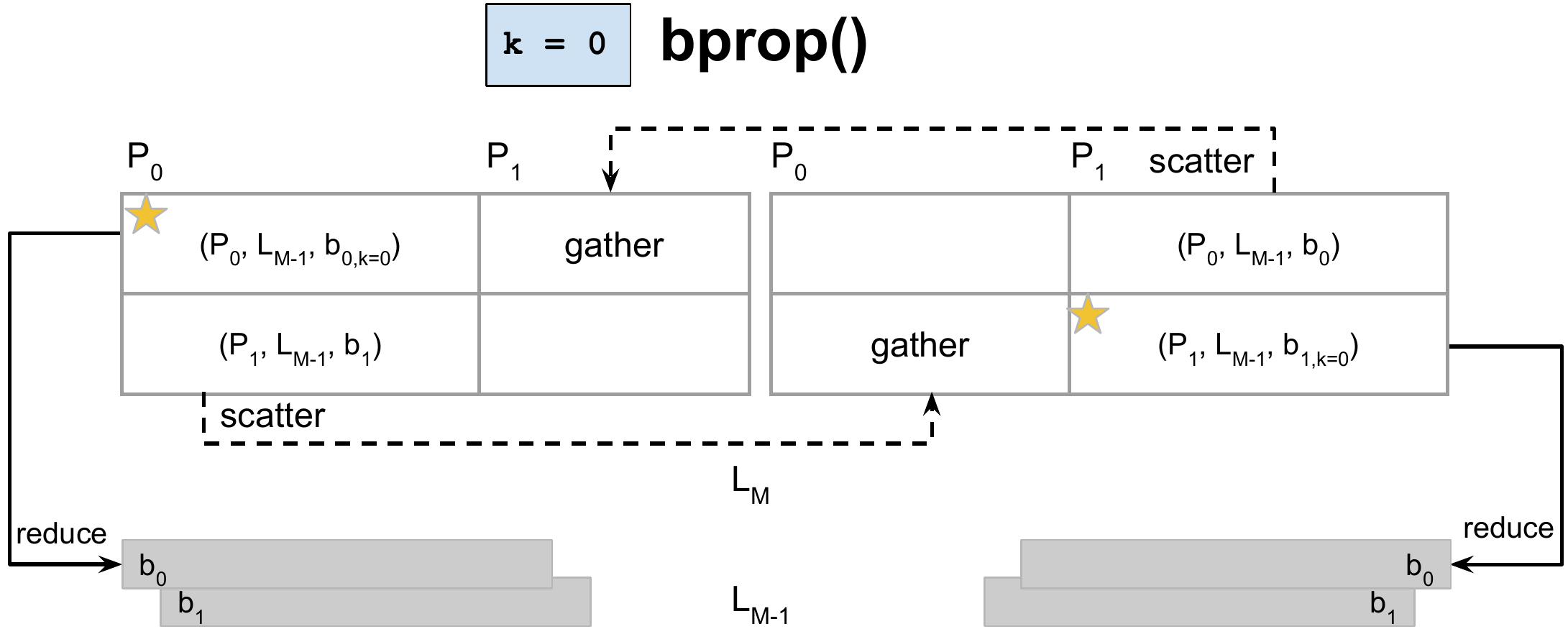}
        \subcaption{Communication of gradients with the other worker to reduce during bprop for $k=0$.}
        \label{fig:modulo-b}
    \end{minipage}
    \hfill
    \begin{minipage}[b]{1.0\linewidth}
        \includegraphics[keepaspectratio,width=\textwidth]{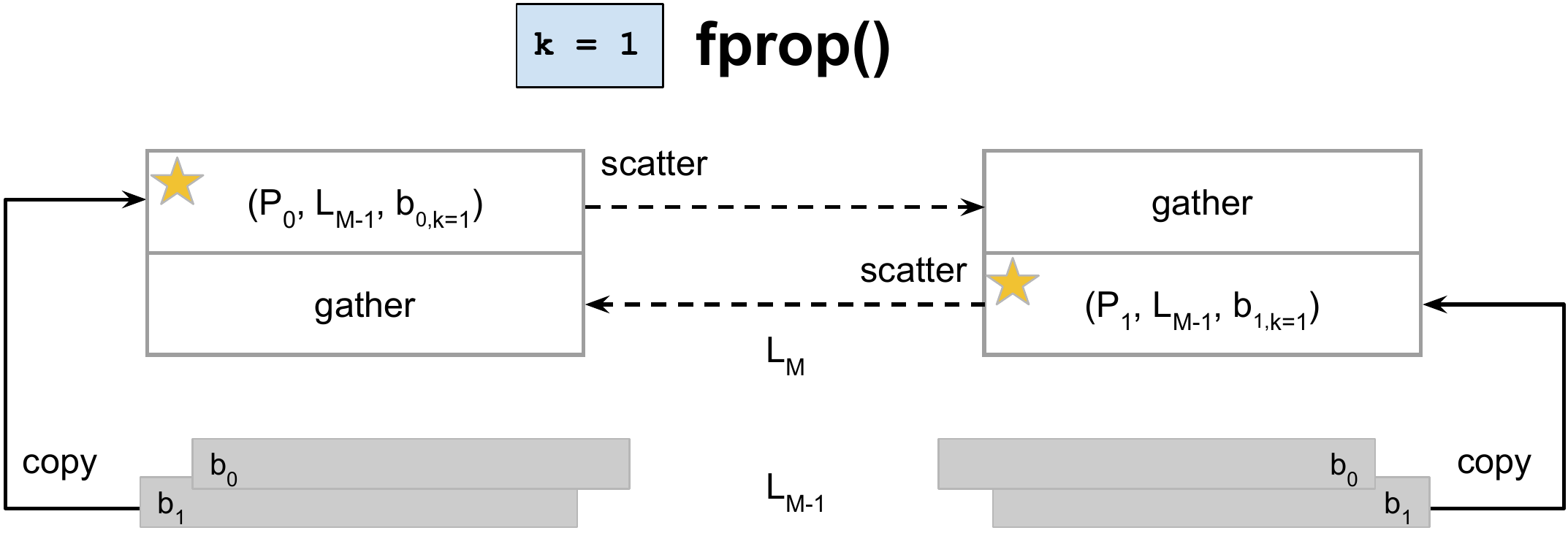}
        \subcaption{Communication of the next $B/K$ local and $B-B/K$ remote batch examples with the other worker during fprop for $k=1$.}
        \label{fig:modulo-c}
    \end{minipage}
    \hfill
    \begin{minipage}[b]{1.0\linewidth} 
        \includegraphics[keepaspectratio,width=\textwidth]{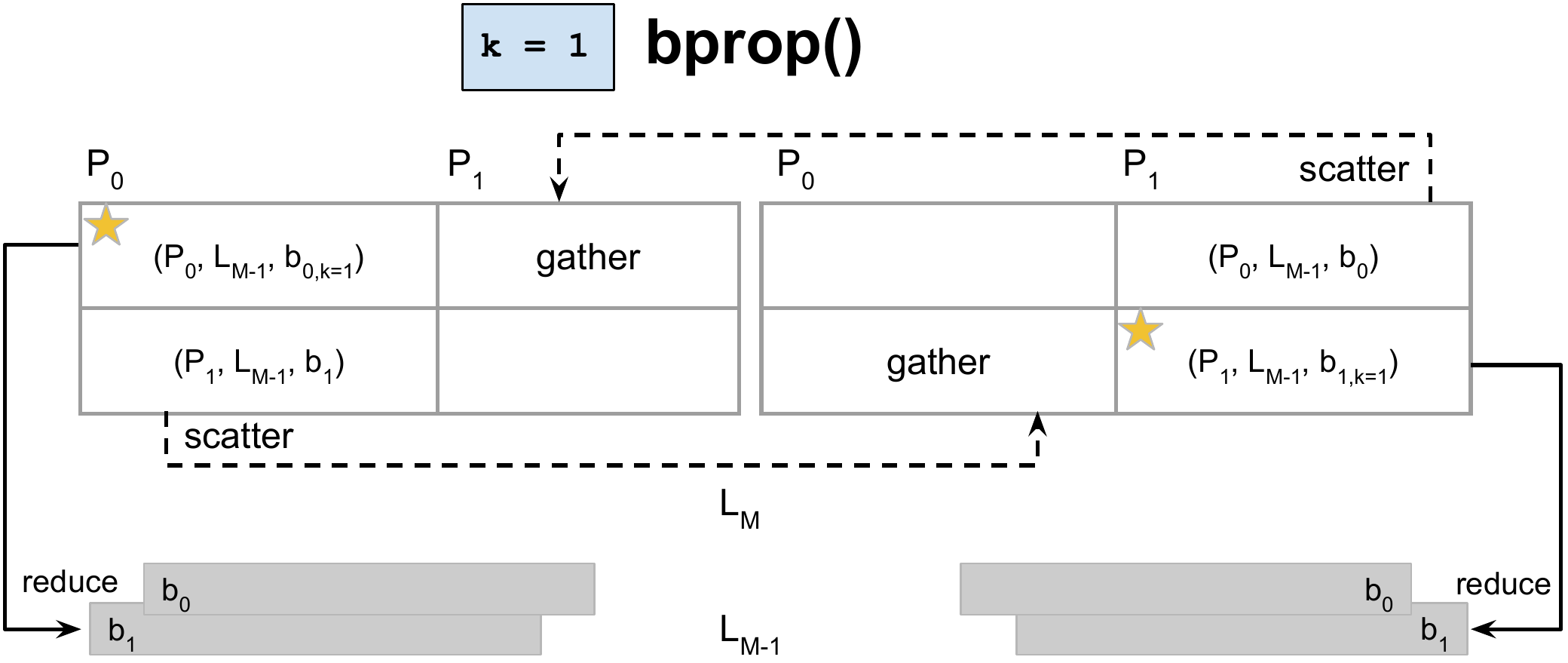}
        \subcaption{Communication of gradients with the other worker to reduce during bprop for $k=1$.}
        \label{fig:modulo-d}
    \end{minipage}
    \caption{\small\bf 
    Modulo layer $\textrm{L}_\textrm{M}$ communication of $B/K$ batch examples in $K=2$ iterations, where $B=2$ and $K=2$ are the batch size and number of workers. 
    $\textrm{L}_\textrm{M-1}$ stands for the layer below $\textrm{L}_\textrm{M}$.
    Each worker is identified by $iProc=0..1$ and responsible for scattering $B/K=1$ local and gathering $B-B/K=1$ remote batch examples with the other per iteration.
    }
    \label{fig:LM}
    \vspace{-0.2in}
\end{figure}

\begin{figure}[ht]
    \begin{minipage}[b]{1.0\linewidth}
        \includegraphics[keepaspectratio,width=\textwidth]{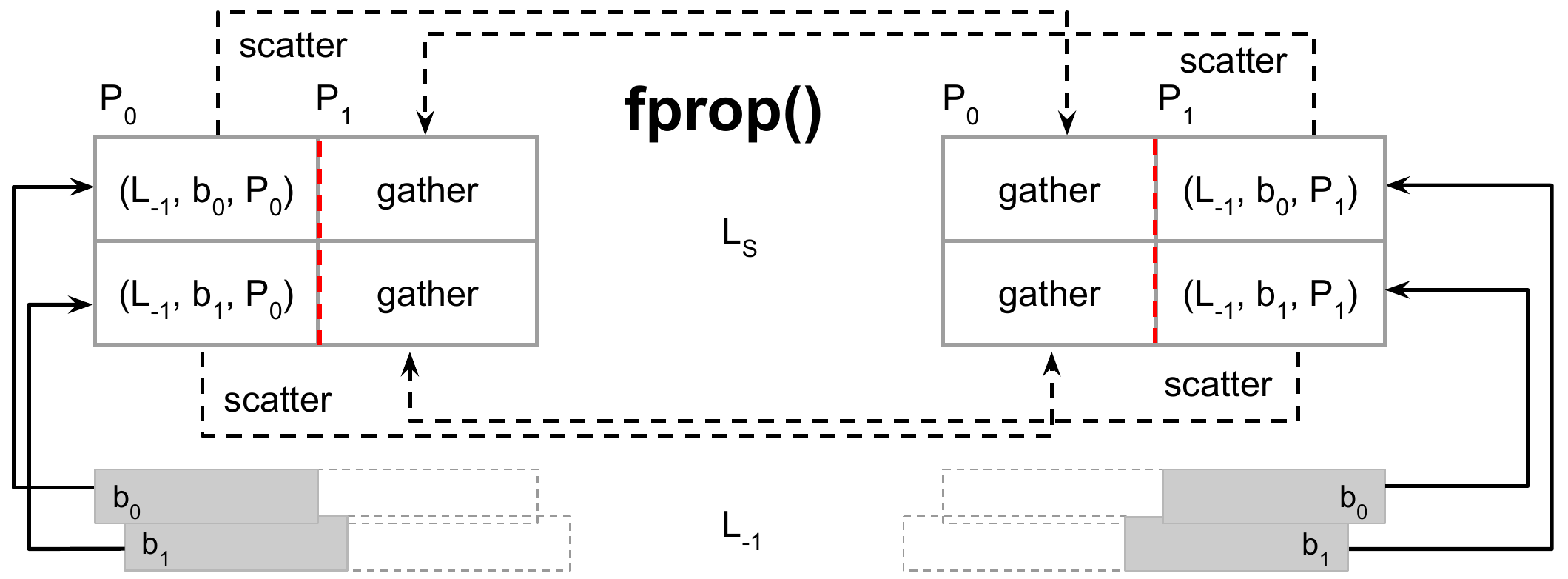}
        \subcaption{Communication of partitioned output with the other worker during fprop.}
        \label{fig:shard-a}
    \end{minipage}
    \hfill
    \begin{minipage}[b]{1.0\linewidth}
        \includegraphics[keepaspectratio,width=\textwidth]{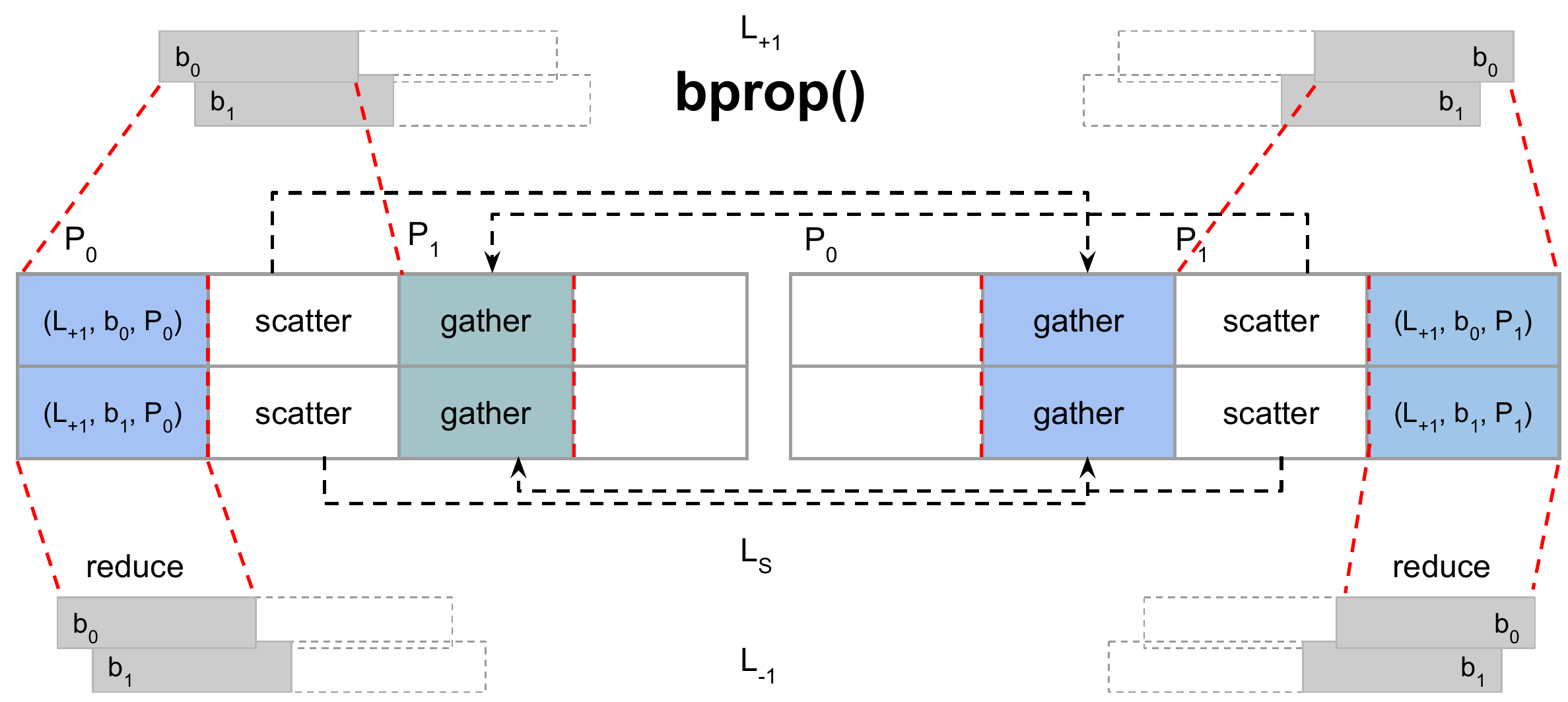}
        \subcaption{Communication of gradients with the other worker to reduce during bprop.}
        \label{fig:shard-b}
    \end{minipage}
    \caption{\small\bf Shard layer $\textrm{L}_\textrm{S}$ communication of partitioned output and gradients. $\textrm{L}_\textrm{+1}$ and $\textrm{L}_\textrm{-1}$ stand for the partitioned layers above and below.}
    \label{fig:LS}
    \vspace{-0.2in}
\end{figure}

\begin{figure}[ht]
    \begin{minipage}[b]{1.0\linewidth}
        \centering
        \includegraphics[keepaspectratio,width=\linewidth]{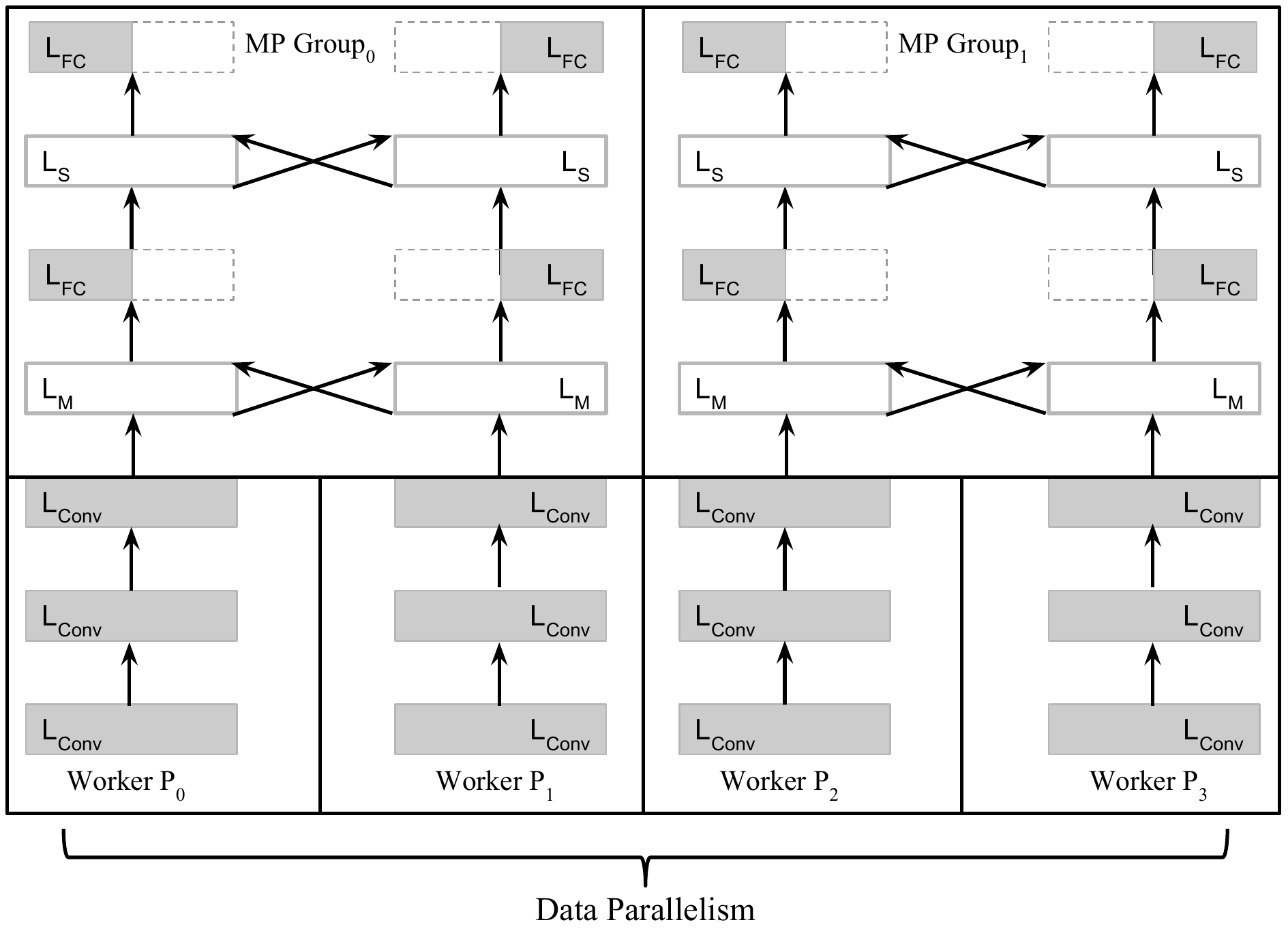}
        \subcaption{Four workers form two MP groups of size two by setting $mp=2$ to enable GMP.}
        \label{fig:mp2-nproc4}
    \end{minipage}
    \hfill
    \begin{minipage}[b]{1.0\linewidth}
        \centering
        \includegraphics[keepaspectratio,width=\linewidth]{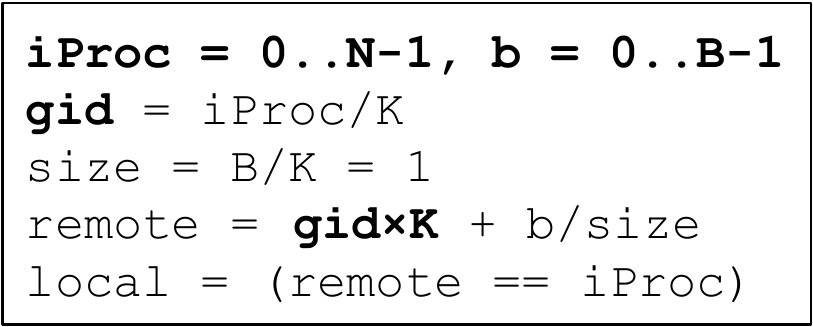}
        \subcaption{GMP batch example mapping to worker id $remote$ with $gid$ to identify the MP group worker $\textrm{P}_i$ belongs to for $\textrm{L}_\textrm{M}$. Assume there are $N$ workers and the MP group size is $K=mp$.}
        \label{fig:modulo-gmp}
    \end{minipage}
    \caption{\small\bf A CNN with GMP enabled is trained on four workers in two MP groups. $\textrm{L}_\textrm{M}$ is extended to support GMP accordingly.}
    \label{fig:gmp}
    \vspace{-0.2in}
\end{figure}

\paragraph{Modulo Layer}

The stateful scheduling for the $K$ iterations is encapsulated in a modulo layer $\textrm{L}_\textrm{M}$ which is transparent in the original CNN.
Each iteration selects a distinct subset of $B/K$ examples of the local batch from the layer below for fprop and the corresponding gradients for bprop based on a modulo $K$ counter $k$.
Specifically in fprop, the $B/K$ local batch examples are broadcast by scattering to the other worker machines and the $B-B/K$ batch examples are gathered back simultaneously.
$\textrm{L}_\textrm{M}$ needs to determine the mapping of the $B/K$ batch examples locally and the others remotely according to the worker id, $iProc=0..K-1$.
Suppose $B$ is a multiple of $K$ such that a worker takes distinct $B/K$ local examples of the batch each iteration.
Clearly, worker $\textrm{P}_i$ can map batch examples $\textrm{b}_{i \times B/K..(i+1) \times B/K - 1}$ locally across iterations.
In general, a batch example is mapped to the worker with id derived from dividing the batch example index by $B/K$.
If the id is equal to worker $\textrm{P}_i$, the batch example must be selected locally.
Otherwise, it must be gathered from some remote worker.

In bprop, the gradients corresponding to the $B/K$ local batch examples are gathered and reduced from the other workers by summing up the values while the gradients of the $B-B/K$ batch examples are broadcast back to the other corresponding workers.
Every $K$ iterations, the gradients of the local batch is fully computed and normal bprop through the convolutional layers resumes to update the parameters.
On the other hand, the FC layer parameters are updated $K$ times more than the convolutional layers.
To compensate the inconsistency, the gradients are divided by $K$ for the FC layers to learn.

Figure \ref{fig:LM} visualizes the fprop and bprop of the modulo layer $\textrm{L}_\textrm{M}$ of a minimal CNN with $K=2$ workers and batch size $B=2$.
Each layer produces output corresponding to the batch examples $\textrm{b}_\textrm{0..1}$.
The starred batch example of $\textrm{L}_\textrm{M}$ serves a local one selected and copied from the previous layer $\textrm{L}_\textrm{M-1}$.
The starred batch example index is determined following the procedure in the box of Figure \ref{fig:modulo-a}.
$remote=b/size$ is the worker id from which the batch example should be derived.
Comparing the id with $iProc$ determines whether to copy locally or gather remotely.
For $k=0$, worker $\textrm{P}_0$ copies $\textrm{b}_0$ from $\textrm{L}_\textrm{M-1}$ to the starred output batch example $\textrm{b}_{0,k=0}$ since $size=B/K=1$ and $remote=b/size=0/1=0$.
The other worker $\textrm{P}_1$ copies $\textrm{b}_0$ from $\textrm{L}_\textrm{M-1}$ to its starred output batch example $\textrm{b}_{1,k=0}$ since $size=B/K=1$ and $remote=b/size=1/1=1$.
Their non-starred output batch examples should be gathered from each other for $remote$ unequal to their $iProc$.
$\textrm{L}_\textrm{M}$ then scatters the starred batch examples and gathers those non-starred ones simultaneously.
Figure \ref{fig:modulo-b} shows the bprop for $k=0$, where the gradients corresponding to the starred batch examples must be gathered from the other workers to reduce in case of loss of information.
If not starred, the gradients are scattered back to the other workers.
Figure \ref{fig:modulo-c} and Figure \ref{fig:modulo-d} illustrate similar fprop and bprop for $k=2$ except that the starred batch example is copied from output $\textrm{b}_1$ of $\textrm{L}_\textrm{M-1}$.


\paragraph{Shard Layer}

With partitioned layer output as input, a shard layer $\textrm{L}_\textrm{S}$ broadcasts by scattering and gathers partial output from the other workers in fprop to feed the full input to the next layer.
In bprop, a partitioned layer computes the gradients w.r.t. the full input to the layer below.
Therefore, the gradients must be scattered back and gathered to reduce from the other workers without loss of information through shard layers.
Specifically, the gradients are composed of a local partition and the other partitions for remote workers.
A worker must gather the gradients corresponding to the local gradient partition from the other workers while scattering the other partitions to the corresponding remote workers.

Figure \ref{fig:LS} demonstrates the fprop and bprop of the $\textrm{L}_\textrm{S}$ of a CNN with $K=2$ workers and batch size $B=2$.
Since $\textrm{L}_\textrm{S}$ only receives $1/K$ output from $\textrm{L}_\textrm{-1}$ the layer below, each worker scatters the local batch partition while gathering the remaining from the others as shown in Figure \ref{fig:shard-a}.
Accordingly, the local batch partition and the remaining to gather are $1/K=1/2$ and $\frac{K-1}{K}=1/2$ of the batch size $B$.
The bprop in Figure \ref{fig:shard-b} illustrates communication of the gradients in terms of the layer partitioning with the other workers because the gradients from the above partitioned layer still cover the full input.
Nonetheless, only $1/K$ of the gradients need to be reduced to pass down.

\subsection{Group MP Extension}

Even though scheme B/K is scalable up to the layer above the modulo layer, it is not the case with shard layers.
To trade off space requirements with communication overhead in a configurable way for MP, \toolshort~introduces the group MP (GMP) extension to address the scalability issue not covered by~\cite{krizhevsky2014one}.
The intuition is to perform DP above the modulo layer in the number of MP groups, each of which performs MP internally.
Hence, the shard layer communication can be limited only within one MP group but inter-group communication becomes necessary for model shard parameter updates.
All the model builder needs is to set the training parameter of MP group size $mp$.
\toolshort~takes care of the distributed model complexity automatically.
We expect \toolshort~with GMP to facilitate configuring hybrid DP and MP in terms of available hardware resources and affordable communication infrastructure.

To make it clear, consider a cluster of four machines.
Pure DP is a special case when $mp=1$, where there is no intra-group communication and the inter-group communication is the same as pure DP.
When $mp=2$, there are two MP groups in data parallel as shown in Figure \ref{fig:mp2-nproc4}.
When $mp=4$, there is only one MP group as proposed by~\cite{krizhevsky2014one}.
Figure \ref{fig:modulo-gmp} illustrates the adaptation for $\textrm{L}_\textrm{M}$ to derive the worker id mapping from a batch example index with GMP enabled given $N$ workers and the MP group size $K=mp$.
The difference is to identify the MP group $gid$ to which the worker belongs and its intra-group offset. 
The evaluation section later demonstrates the trade-off between space requirements and communication overhead.

%% file: sections/impl.tex
\section{Implementation}\label{sec:impl}

\toolshort~implements a hybrid data and model parallel solution in C++ to train CNNs with SGD in mini-batches.
Training through convolutional layers allows for DP in the number of total worker machines.
Each worker trains a model replica and exchanges the full set of parameters up to the modular layer periodically in the specified number of communication batches.
Training through FC layers involves MP with DP in the number of MP groups.
Each worker trains a model shard within its MP group by communicating the output of the FC layers while exchanging the model shard parameters for model averaging across MP groups.
On the other hand, \toolshort~abstracts the communication complexities between workers under the hood through the global address space programming interface (GASPI) to access the RDMA infrastructure such as Ethernet or InfinitBand (IB).
The GASPI implementation is based on GPI-2 \cite{Grunewald2013}.
For ease of development, we provide Torch \cite{collobert2011torch7} like CNN construction through Lua bindings.
CNN builders are allowed to specify the MP group size, batch size, model averaging period in the number of batches, and communication graph in a peer-to-peer or parameter server fashion that applies to inter-group MP model shard averaging as DP.

%% file: sections/evaluation.tex
\section{Evaluation} 
\label{sec:evaluation}

\begin{table}[tbp]
\centering
\begin{tabular}{|c||c|r|c|}
 \hline
{\bf Layer} & 
{\bf I/O Dimension} & 
{\bf Parameters} & 
{\bf \%}                     \\
\hline
Conv0 & 32x64         & 1728       & \multirow{7}{*}{24.83} \\
Conv1 & 64x64         & 36864      &                        \\
Conv2 & 64x128        & 73728      &                        \\
Conv3 & 128x128       & 147456     &                        \\
Conv4 & 128x256       & 294912     &                        \\
Conv5 & 256x256       & 589824     &                        \\
Conv6 & 256x256       & 589824     &                        \\
\hline
FC0   & 4096x1024     & 4194304    & \multirow{3}{*}{75.17} \\
FC1   & 1024x1024     & 1048576    &                        \\
FC2   & 1024x10       & 10240      &
\tabularnewline \hline
\end{tabular}
\caption{Layer-wise parameters of the VGG variant. The I/O dimensions of convolutional layers are simplified by channels.}
\label{table:params}
\end{table}

In this section, we evaluate \toolshort~by training a VGG variant~\cite{simonyan2014very} over the CIFAR-10 ~\cite{cifar10:web} dataset.

\subsection{Experimental Setup}

The VGG variant consists of 11 layers with 7.5M parameters in total.
Table \ref{table:params} shows input/output dimensions and the number of parameters of those trainable layers.
Clearly, the three FC layers consume more than 75\% parameter memory usage.
This implies significant memory savings may be achieved by splitting those FC layers with \toolshort.

We conduct all the experiments on a research cluster of up to 32 machines connected via an infiniBand backplane.
Each machine is equipped with 64GB DDR3 DRAM, and an Intel Xeon 8-core 2.2GHz Ivy-Bridge processor that supports SSE 4.2/AVX. 
The cluster of machines are connected via Mellanox Connect-V3 56Gbps InfiniBand cards. 
Our InfiniBand infrastructure provides a peak throughput slightly over 40Gbps after accounting for the bit-encoding overhead for reliable transmission. 
All machines load the input dataset from a shared NFS partition. 
All reported times do not account for the initial one-time cost for loading the dataset into memory. 
All times are reported in seconds.

\subsection{Performance Results}

Next, we present the evaluation results of \toolshort\ to answer the following questions:
\begin{itemize}
\item How does the performance scale out despite MP?
\item What is the communication overhead that MP may introduce?
\item What is the trade-off between throughput and memory usage with \toolshort?
\end{itemize}

\begin{figure}
    \begin{minipage}[b]{1.0\linewidth}
        \includegraphics[keepaspectratio,width=\textwidth]{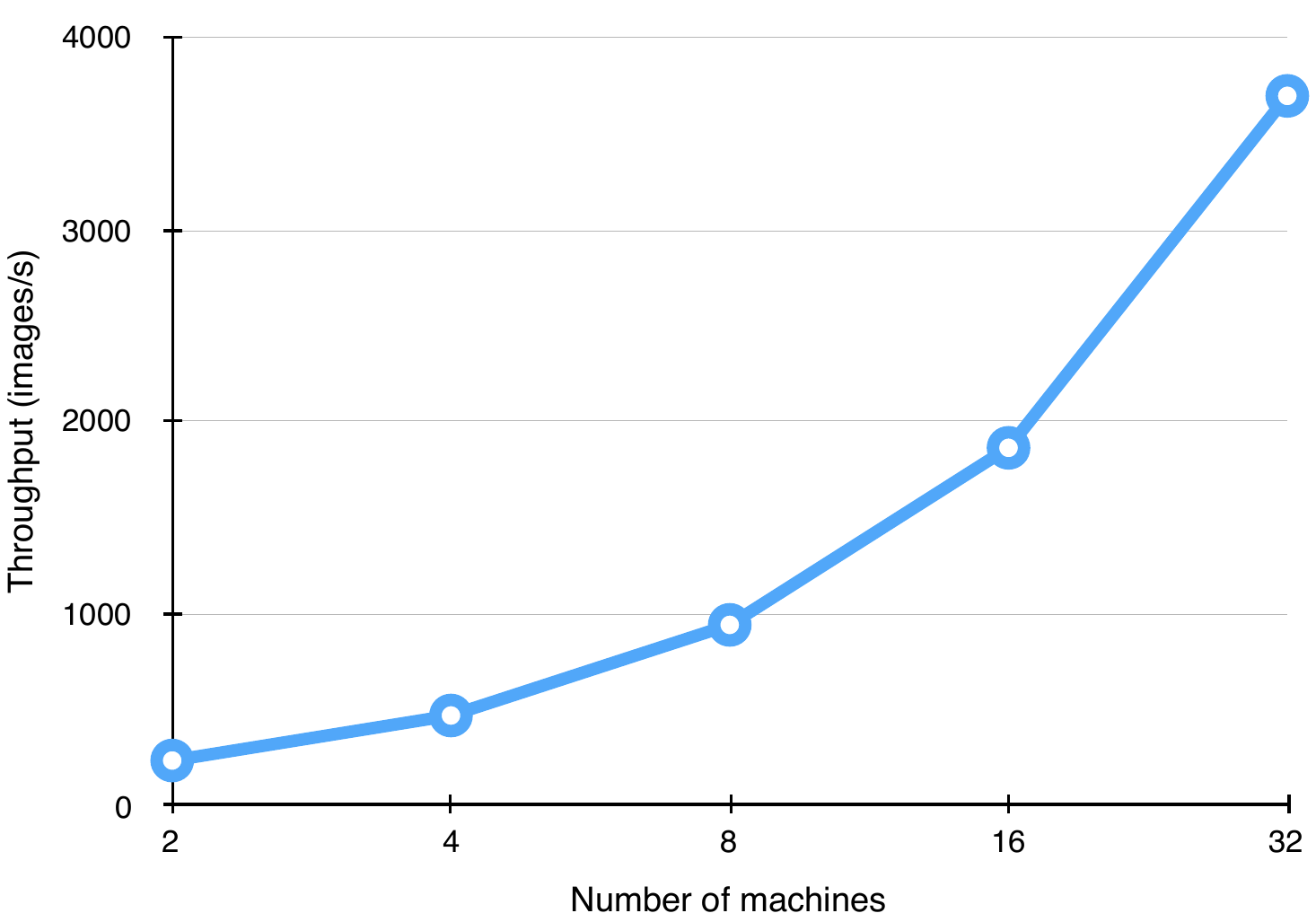}
        \subcaption{Nearly linear scaling of throughput for $MP=2$ with different numbers of machines.}
        \label{fig:scaling}
    \end{minipage}
    \hfill
    \begin{minipage}[b]{1.0\linewidth}
        \includegraphics[keepaspectratio,width=\textwidth]{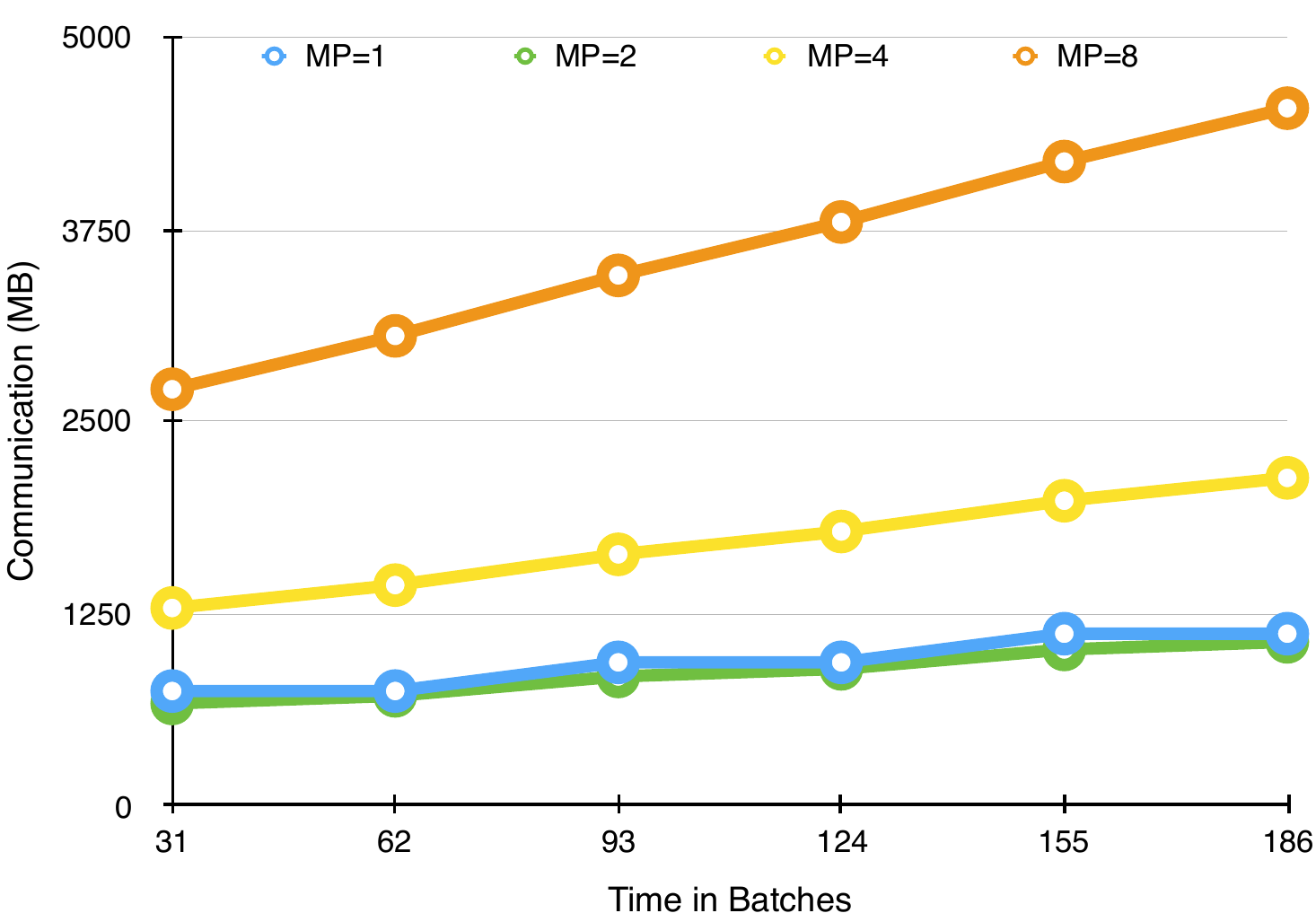}
        \subcaption{Communication overhead w.r.t. different MP group size on a cluster of eight machines.}
        \label{fig:commMP}
    \end{minipage}
    \hfill
    \begin{minipage}[b]{1.0\linewidth}
        \includegraphics[keepaspectratio,width=\textwidth]{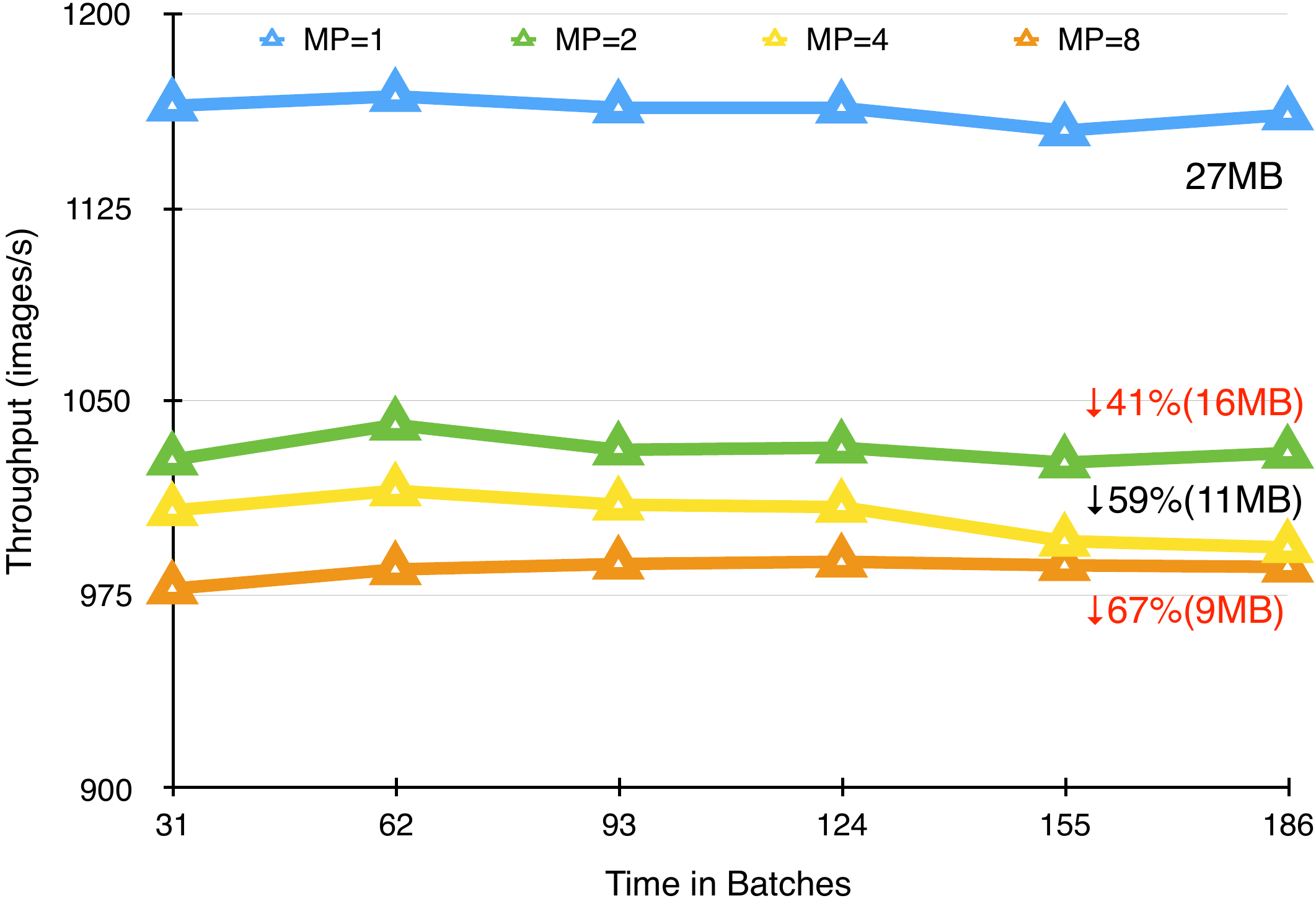}
        \subcaption{Throughput trade-off with memory usage of different MP group size on a cluster of eight machines.}
        \label{fig:throughputMP}
    \end{minipage}
    \caption{\small\bf SplitBrain throughput scaling, communication overhead, and trade-off with memory usage.}
    \label{fig:performance}
    \vspace{-0.2in}
\end{figure}

While the full performance numbers are listed in table \ref{table:throughput}, we highlight the throughput scaling, communication overhead, and trade-off with memory usage on a cluster of 8 machines in figure \ref{fig:performance}.
Figure \ref{fig:scaling} shows the throughput scaling with different numbers of machines for MP group size 2 is nearly linear, indicating \toolshort~performance remains benefiting from DP despite potential MP communication overhead.
Figure \ref{fig:commMP} demonstrates the communication overhead w.r.t. different MP group sizes.
Larger MP group size increases communication overhead drastically but the communication for DP is reduced for fewer parameters to exchange.
This is because MP communication is required for training each example whereas DP may exchange the parameters in some longer period.
When MP group size is 2, the communication overhead is comparable to pure DP.
Nonetheless, the synchronization overhead for training each example inevitably slows down the throughput.
What is implied is that the proposed GMP extension is essential to limit the MP communication overhead in practice.
Figure \ref{fig:throughputMP} illustrates the trade-off between throughput and parameter memory usage.
Pure DP as the baseline is supposed to achieve the highest performance with the most memory usage.
Previous work \cite{krizhevsky2014one} realized MP over all the worker machines attains the lowest performance with the most memory savings.
With GMP, \toolshort~on the other hand provides configurable performance trade-off with memory usage while delivering higher performance than \cite{krizhevsky2014one}.

\begin{table}[tbp]
\center
\footnotesize
\begin{tabular}{|c||l|c|c|r|}
\hline
\centering {\bf Machines} &
\centering {\bf Dataset} &
\centering {\bf DP} &
\centering {\bf MP} &
\centering {\bf images/sec}
\tabularnewline  \hline
1  &  CIFAR-10   & 1   & 1 & 121.99 \tabularnewline \hline
\hline
2  &  CIFAR-10   & 2   & 1 & 247.43 \tabularnewline \hline
2  &  CIFAR-10   & 1   & 2 & 235.72 \tabularnewline \hline
\hline
4  &  CIFAR-10   & 4   & 1 & 489.62 \tabularnewline \hline
4  &  CIFAR-10   & 2   & 2 & 470.1 \tabularnewline \hline
4  &  CIFAR-10   & 1   & 4 & 421 \tabularnewline \hline
\hline
8  &  CIFAR-10   & 8   & 1 & 965.92 \tabularnewline \hline
8  &  CIFAR-10   & 4   & 2 & 941.84 \tabularnewline \hline
8  &  CIFAR-10   & 1   & 8 & 520 \tabularnewline \hline
\hline
16  &  CIFAR-10   & 16   & 1 & 1946.99 	    \tabularnewline \hline
16  &  CIFAR-10   & 8    & 2 & 1863.5     \tabularnewline \hline
\hline
32  &  CIFAR-10   & 8   & 8  & 2062.84   	\tabularnewline \hline
32  &  CIFAR-10   & 8   & 4  & 3293.68    \tabularnewline \hline
32  &  CIFAR-10	  & 16  & 2  & 3695.64   \tabularnewline \hline
32  &  CIFAR-10   & 32  & 1  & 3896.27    	\tabularnewline \hline
\end{tabular}
\caption{\small \bf CIFAR-10 throughputs in combinations of DP and MP}
\label{table:throughput}
\vspace{-0.2in}
\end{table}




%% file: sections/related.tex
\section{Related Work}
\label{sec:related}
Our work is inspired by existing work on distributed ML and hybrid parallelism.

\paragraph{Distributed ML}
Scalability and convergence guarantee are the keys to distributed ML.
The former aims at reducing the communication complexity and synchronization overhead for scaling out.
The latter specifies the conditions that distributed ML algorithms or frameworks must hold to ensure the correctness.
DistBelief~\cite{Dean2012-kv}, Project Adam~\cite{chilimbi2014adam} and the parameter server~\cite{Li2014-vr} use master-client communication to train model replicas in DP.
However, their approaches scale poorly as the master or parameter server becomes the bottleneck.
Distributed ML frameworks such as MATL~\cite{li2015malt} and Horovod~\cite{Sergeev2018-xc} propose generalized communication APIs including $scatter$ and $gather$ to hide the underlying communication complexities.
When implementing the broadcast of \toolshort~using their APIs, the underlying communication may follow a peer-to-peer pattern in a Halton sequence or a ring, leading to significant speedup and bandwidth savings as well as fast failure recovery.
In terms of convergence guarantee, bulk synchronous (BSP)~\cite{Valiant1990-ff} , asynchronous lock-free~\cite{recht2011hogwild} and Stale Synchronous Parallel (SSP)~\cite{ho2013more} solutions are proposed with proof of correctness for aforementioned distributed ML frameworks to apply.
For simplicity and efficiency, \toolshort~assumes BSP communication within MP groups across FC layers is supported for MP while DP is up to the underlying distributed ML framework.

\paragraph{Hybrid Parallelism}
In~\cite{krizhevsky2014one}, three MP schemes are proposed but only the second one is evaluated.
Built on the third scheme, our \toolshort~further introduces the GMP extension to trade off communication efficiency for reduced memory requirements with configurable hyperparameters.
On the other hand, recent work on scaling large language models~\cite{wu2016google} also proposes a combination of DP and MP.
Specifically, their MP is applied to the encoder and decoder LSTM in the architecture by parallelizing each layer on a different GPU.
In contrast, our work focuses on CNNs and automates the partitioning of each FC layer in a scalable way across multiple machines within groups. 




%% file: sections/conclusion.tex
\section{Conclusion} \label{sec:conclusion} 

In this paper, we present a systematic transformation of CNNs with configurable hybrid parallelism for distributed machine learning.
Two communication constructs, the modulo and shard layers, are proposed to automate the transformation while hiding the complexities from the programmer.
The configurability comes from the GMP extension that trades off between communication overhead and memory usage.
The evaluation shows that with a small MP group size, the speedup is nearly linear.
Besides, the sweet spot between pure DP and different MP group sizes can now be explored, which was unavailable in previous work.

In the future, we plan to investigate the design space of fine-grained model partitioning given a resource budget.
Further, considering the dynamic workload of model training and inference over time requested by multiple users with different priorities, the optimal partitioning may require adaptive migration for the best trade-off w.r.t. the budget and performance constraints.